\documentclass[
	12pt,				% tamanho da fonte
	openright,			% capítulos começam em pág ímpar (insere página vazia caso preciso)
	oneside,			% para impressão em verso e anverso. Oposto a oneside
	a4paper,			% tamanho do papel. 
	]{article}

\usepackage{arxiv}

\usepackage{lmodern}

\usepackage{indentfirst}	% Indenta o primeiro parágrafo de cada seção.
\usepackage[utf8]{inputenc} % allow utf-8 input
\usepackage[T1]{fontenc}    % use 8-bit T1 fonts
\usepackage{hyperref}
\usepackage{url}            % simple URL typesetting
\usepackage{booktabs}       % professional-quality tables
\usepackage{amsfonts}       % blackboard math symbols
\usepackage{nicefrac}       % compact symbols for 1/2, etc.
\usepackage{microtype}      % microtypography
\usepackage{lipsum}
\usepackage{caption}
\usepackage{graphicx}			% Inclusão de gráficos
\usepackage{xcolor}
\usepackage{listings}
\usepackage{floatrow}

\definecolor{codegreen}{rgb}{0,1,0}
\definecolor{codegray}{rgb}{0.5,0.5,0.5}
\definecolor{codepurple}{rgb}{0.58,0,0.82}
\definecolor{backcolour}{rgb}{0.95,0.95,0.92}

\title{An Empirical Accuracy Law for Sequential Machine Translation: the Case of Google Translate}

\author{
    Lucas N.~Sequeira \\
    \texttt{lucasnseq@usp.br} 
\And
    Bruno S.~Moreschi \\
    \texttt{brunomoreschi@gmail.com} \\
\And
    Fabio G.~Cozman \\
    \texttt{fgcozman@usp.br} 
\And
    Bernardo~Fontes \\
    \texttt{bernardoxhc@gmail.com} 
}

\begin{document}
\maketitle

\begin{abstract}
In this research, we have established, through empirical testing, a law (Accuracy Empirical Law - AEL) that relates the number of translating hops to translation\textbf{}
accuracy in sequential machine translation with Google
Translate. Both accuracy and size decrease with the number
of hops; the former displays a decrease closely
following a power law.
Such a law allows one to predict the behavior of translation
chains that may be built as society increasingly depends on
automated devices. 
%
%Translation is an essential and old human endeavor; its automation is now a common service. %Google Translate, the object of study here, is a widely used but rather opaque translation service. 
%We report here an investigation on sequential machine translation (focused on Google translate), confirming some known facts but also identifying new ones. In particular, we describe the behavior of text size variation and how one can capture accuracy of sequential translation by a mathematical law. %Our results allow a broader understanding about machine's behavior and is a reference for further studies in this area.
\end{abstract}

% keywords can be removed
% \keywords{Machine Translate \and Accuracy Behavior \and Google}

\section{Introduction}

\begin{figure}[t]
    \caption{The number of languages available in free machine translation web services.}
    \label{fig:services}
    \includegraphics[width=14cm]{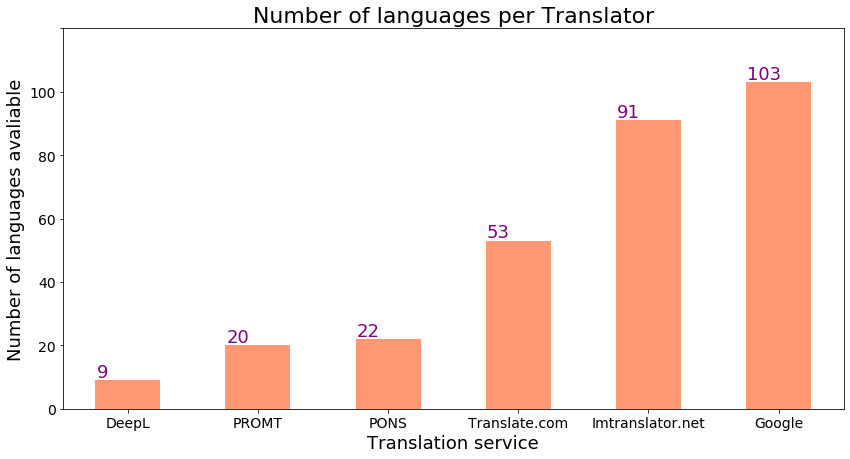}
\end{figure}

%Translation is an old human activity with mixeds on ancient commercial trade \cite{kwintessential_2018}. Indeed,
Some of the first translations turned Sumerian poems to Asian languages about 4500 years ago \cite{alster_1995, hutchins_1995}. In 1440, with the development of the printing press, it became easier to circulate texts around; consequently, translation services became more common. 
Since the popularization of the internet we live in a world where translation is essential
%%%
%\paragraph{}Much discussed on Linguistic field, the history of translation walks side by side with the history of the first human interactions, in such contexts like survival, effectiveness and commercial trades \cite{kwintessential_2018}. Some of the first texts which have been translate, were the Sumerian poems translated into several Asiatic languages and date 4500 years ago, in Mesopotamian age \cite{alster_1995, hutchins_1995}. In 1440, with the development of the printing press, it became easier to circulate texts around and, consequently, the translation practice intensified. Since the popularization of the internet from the end of the 20th century, today we probably live the greatest mark in this field. Gutenberg's press gave way to increasingly powerful computers and digital systems.
%%
with effective machine translation systems. 
Starting from the early 40's where the first machine translators were conceived
\cite{cheragui_2012}, to various periods where rule-based systems were popular,
we now have dominance of data-based schemes. 
%According to Chéragui \cite{cheragui_2012}, the history of machine translation is divided into five periods: the early 40's and 50's with the first machine translator (Georgetown University and IBM); the formalization of this subfield with events such as the International Machine Translation Conference (1961); then, two periods of emergence of rules-based systems such as REVERSO (Russia), SYSTRAN1 (USA), DUET and ATLAS2 (Japan); and, in the last period, the fusion of rule models with probabilistic and example-based models. 
Deep learning models, such as Google Translate's Neural Machine
Translation, now  command this quickly moving field (Figure \ref{fig:services}) \cite{mcguire_2019}.
 In this process, translation techniques have
become so intricate that a complete understanding of its constituent
elements is not feasible anymore. 

In this paper we wish to understand features of multi-language machine translation systems. We are interested in sequential translation, which we could briefly describe as a set of step by step translations of a single text into a chain of different languages, as if machines were to play a kind of ``Chinese whispers'' game.\footnote{Sometimes called   ``telephone'' or ``wireless phone'', this is a game where people stay next to each other forming a line, the first person in line whispers to the next one a sentence, and as the next person hears the sentence, whispers to the next in line what has been understood and so on. The goal is to see what happens with the input sentence along the chain of whispers.} That is, an input sentence is to be sequentially translated into different languages in several steps, and our goal is to see what happens with the output texts. Imagine this happening in a world where many machines operate in various languages and translation may occur in several hops until it reaches a particular agent. What is this agent to expect from the receiving message as compared to the original one?

As will be demonstrated in the paper, 
translation accuracy measured by the Google Bilingual Evaluation Understudy (GLEU) score decreases according to a
power law with the number of translation hops. Remarkably, the
size of translated texts also decreases monotonically with 
the number of translation hops, even though a simple mathematical
relation does not seem to emerge.
We used Google Translate with chains of up to 284 translation hops,
relating 71 distinct languages and investigating hundreds of
different translation orders. 
These results open a new perspective on how to attain an holistic understanding of 
machine translation, as they show that empirical laws can
capture features of this rapidly evolving field. 

In Section \ref{sec:metodologia} we describe the methodology we followed in our experiences. Section \ref{sec:resultados} carries our results: we describe the variation of text size and translation accuracy as a function of the number of sequential translations applied to a text. In Section \ref{sec:conclusoes} we raise points that may direct future research.

%\paragraph{}For that, in section \ref{sec:metodologia}, we describe the methodology of the research. In section \ref{sec:resultados}, we present the main results obtained, which are related to variation of the text size among sequential translation and measurement of accuracy in the Google Translate. Finally, in section \ref{sec:conclusoes}, we conclude by raising points that may serve for future research in the topic.\footnote{This research would not be possible without the collaboration of Pedro Barbosa, Paloma F. Victor (for unequivocally supporting Sequeira in the analysis of the results), members of GECID+AIA, employees of Inova USP and PIBIQ (Institutional Scientific Initiation Scholarship Program). Thank you so much for your trust and support.}

\section{Methodology}
\label{sec:metodologia}

As we are interested in sequential machine translation, we started by developing a script that employs the Google Translate API  \cite{google_2019} so as to produce chains of translations.
We selected 71 languages amongst the 103 languages available in the service; those 71 languages correspond to the official languages \cite{wikipediaLANG_2019} of the countries listed in the Global Human Development Index (HDI) list \cite{pnud_brasil_2015}. 

We define {\em translation chain} as a sequence of translation hops, each hop corresponding to a call to the Google Translate API — a text is sent in one language and translated to the next language in the chain order. 

We  selected four texts to start the translation chains, each chain consisting of 284 translation hops. 
The first text is a short excerpt from a Master thesis \cite{bueno_2011}
about the Argentine author Jorge Luis Borges\footnote{Jorge Francisco Isidoro Luis Borges Acevedo (24 August 1899 – 14 June 1986) was an Argentine short-story writer, essayist, poet and translator, and a key figure in Spanish-language and universal literature.}. 
The text discusses the translation of literary texts, in particular
the book \textit{One Thousand and One Nights}.\footnote{One Thousand and One Nights is a storybook of unknown authorship told from generation to generation that to this day goes through several translations \cite{mundo_estranho_2018, britannica_2019}.} 
We use t3 to refer to the 
following text in English:

\fbox{%
  \parbox{\textwidth}{
    {\fontfamily{qcr}\selectfont
    "For him, each reading of any text will always provide a new resizing and understanding of that text. Metaphorically speaking, he positions himself before the texts as the bather of the river of Heráclito, in which it is impossible to enter twice because his course is constantly changing."}
  }%
}

We use t1 to refer to the Portuguese version of the same text:

\fbox{%
  \parbox{\textwidth}{
    {\fontfamily{qcr}\selectfont
    "Para ele, cada leitura de qualquer texto sempre proporcionará um novo redimensionamento e entendimento desse texto. Metaforicamente falando, ele se posiciona diante dos textos como o banhista do rio de Heráclito, no qual é impossível entrar duas vezes devido a seu curso estar em constante mutação."}
  }%
}\\
 
We use t4 to refer to a 258-word excerpt from 
the book One Thousand and One Nights in its English version:

\fbox{%
  \parbox{\textwidth}{
    {\fontfamily{qcr}\selectfont
    "The Khaleefeh Hároon Er-Rasheed had gone forth this night to see and hear what news he could collect, accompanied by Jaafar his Wezeer, and Mesroor his executioner. It was his custom to disguise himself in the attire of a merchant; and this night, as he went through the city, he happened to pass, with his attendants, by the house of these ladies, and hearing the sounds of the musical instruments, he said to Jaafar, I have a desire to enter this house, and to see who is giving this concert. They are a party who have become intoxicated, replied Jaafar, and I fear that we may experience some ill usage from them; but the Khaleefeh said, We must enter, and I would that thou devise some stratagem by which we may obtain admission to the inmates. Jaafar therefore answered, I hear and obey: and he advanced, and knocked at the door; and when the portress came and opened the door, he said to her, My mistress, we are merchants from Tabareeyeh, and have been in Baghdád ten days; we have brought with us merchandise, and taken lodgings in a Khán; and a merchant invited us to an entertainment this night: accordingly, we went to his house, and he placed food before us, and we ate, and sat awhile drinking together, after which he gave us leave to depart; and going out in the dark, and being strangers, we missed our way to the Khán: we trust, therefore, in your generosity that you will admit us to pass the night in your house; by doing which you will obtain a reward in heaven."}
  }%
}

And finally we use t2 to refer to a 296-word excerpt of the 
same book in its Portuguese version:

\fbox{%
  \parbox{\textwidth}{
    {\fontfamily{qcr}\selectfont
    "Fora, portanto, Djafar, o grão-vizir, quem havia batido à porta das damas por ordem do califa, que não desejava ser reconhecido. Safia abriu; e o vizir, notando, à luz de uma vela que ela trazia, que se tratava de dama de grande beleza, representou perfeitamente bem o seu papel. Fazendo profunda reverência, disse-lhe respeitosamente: “Senhora, somos três mercadores de Mussul, chegados há dez dias, com ricas mercadorias que depositamos num Khan (10), onde nos hospedamos. Estivemos hoje na casa de um mercador desta cidade que nos convidou a visitá-lo. Ofereceu-nos ele excelente refeição; e, como o vinho nos houvesse posto de bom humor, mandou chamar um bando de dançarinas. Já era noite, os músicos tocavam, as bailarinas dançavam e o grupo fazia enorme bulha, quanto a ronda, passando por lá, ordenou que abrissem. Alguns foram detidos. Quanto a nós, fomos bastante felizes e salvamo-nos saltando por sobre um muro; mas, acrescentou o vizir, como somos estrangeiros, e além disso um pouco dominados pelo vinho, tememos encontrar outra patrulha, ou a mesma, antes de chegarmos ao nosso Khan, bem longe daqui. E a ele chegaríamos até inutilmente, pois a porta está fechada, e só será aberta amanhã cedo, apesar de tudo quanto possa suceder. E por isso, senhora, que, tendo ouvido instrumentos e vozes, julgamos que nos seria permitido bater, para vos suplicar abrigo até o romper do dia. Se vos parecermos dignos de participar do vosso divertimento, esforçar-nos-emos por contribuir como pudermos, a fim de recompensarmos a interrupção por nós causada; se não, concedei-nos pelo menos a graça de nos permitir passar a noite no vosso vestíbulo."}
  }%
}\\

%We consider these two starting texts in Portuguese, as we are a research group located in São Paulo, Brazil.

%In Section \ref{subsec:resultadosA} we used three different translation chains (rand1, rand2 and rand3), all with the same 71 languages, but randomly varying the order of the languages and keeping the same frequency for each one. This set of experiments enabled us to see how machine translation would behave along a large random sequence of languages.

In Section \ref{subsec:resultadosA} we analyze the impact of text size variation along sequential machine translation. To do so, first we used three different translation chains (rand1, rand2 and rand3), all with the same 71 languages, but randomly varying the order of the languages.
%and keeping the same frequency for each one. 

\begin{figure}[H]
    \floatbox[{\capbeside\thisfloatsetup{capbesideposition={left,top},capbesidewidth=6cm}}]{figure}[\FBwidth]
    {\caption{A small clip of the Language Family Tree \cite{okrent_2014}. We have two main language families here: Slavic and Romance (Neolatine). In this clip, Slavic could provide a set of common languages using for example Polish, Russian and Ukranian. But to build a mixed language set, we would take a Slavic and a Romance language. %While for the \textit{mixed} group each family could provide a single language for the same set, such as Russian from Slavic family and Romanian from Romance.
    }
     \label{fig:tree_cut}}
    {\includegraphics[width=6cm]{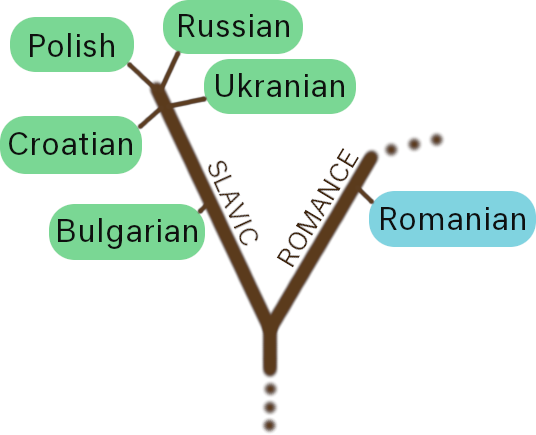}}
\end{figure}

Secondly, we used four different chains, which were divided into two groups: \textit{common} and \textit{mixed}. A common-chain is a sequence of hops that have languages with close common ancestry, such as Portuguese and Italian (both are closely correlated Neolatine languages). A mixed-chain is a sequence of languages that are poorly correlated with each other, such as Russian (Slavic) and Romanian (Romance). Figure \ref{fig:tree_cut} illustrates these differences.

Each of the two common-chains has a set of six nearby languages; nearby languages are languages with recent common ancestry such as Slavic languages (Figure \ref{fig:tree_cut}). Distant languages instead have non-recent common ancestry in the language tree with each other as Sino-Tibetan and Neolatine languages \cite{jingsheng_2008,okrent_2014}; we also take  English always as reference. The first chain of the \textit{common} group has Neolatine languages (Catalan, French, Italian, Portuguese, Romanian and Spanish --- indexed by com1), while the second, Germanic languages (Afrikaans, Danish, Dutch, German, Norwegian and Swedish - indexed by com2).

%In Section \ref{subsec:resultadosB}, we used four different chains, which were divided into two groups: \textit{common} and \textit{mixed}\footnote{\textit{common} is a group of sets that have languages with close common ancestry, such as Portuguese and Italian, both are closely correlated Neolatine languages. \textit{mixed} is a group whose language sets in this group have languages that are poorly correlated with each other, such as Chinese (Sino-Tibetan) and English (Germanic). Figure \ref{fig:tree_cut} illustrates this classification.}. For the \textit{common} group, each of the two chains has a set of nearby languages\footnote{Nearby languages are languages with more recent common ancestry such as Slavic languages (figure \ref{fig:tree_cut}), and distant languages, which have non-recent common ancestry with each other as Sino-Tibetan and Neolatine languages \cite{jingsheng_2008}.} \cite{okrent_2014} and English as reference. The first chain of the \textit{common} group has Neolatine languages (Catalan, French, Italian, Portuguese, Romanian and Spanish - indexed by com1), while the second, Germanic languages (Afrikaans, Danish, Dutch, German, Norwegian and Swedish - indexed by com2).

The procedure was similar for the two mixed-chains, with the difference that each one of both chains (mix1 and mix2) always cluster languages from six different family origins --– Germanic, Indic, Iranian, Neolatine, Sino-Tibetan and Slavic \cite{okrent_2014, stedt_2013, jingsheng_2008} (and English as reference – totalizing seven languages). These language clusters allowed us to study various translation characteristics, from chains with a huge list of languages to chains with small subsets with shared linguistic properties.

Even though language distance must be measured by taking into account factors such as diachronic linguistics, connections with additional languages, language-based conflicts and the effects of language differences on trade~\cite{crowley_bowern_2010}, here we simplify matters by focusing only on  diachronic linguistics, assuming that this simplification captures the impact of machine translation over different language families.

In Section \ref{subsec:resultadosB} we study the accuracy of sequential machine translation  using translation chains discussed in Section \ref{subsec:resultadosA}. We used the GLEU (Google Bilingual Evaluation) as performance metric. Used as an alternative to the BLEU metric in 2016 by Google researchers, this index basically measures how accurate is a translation. It consists of taking the minimum recall and precision value of n-gram pairs between the compared texts; it is  explained in detail in Ref.\ \cite{wu2016googles}. 
We compute the GLEU between the $t$-th text (in English) and the initial text (also in English), a method that we refer to as ``accumulated GLEU''. 

%Besides that, on each step of the sequential translations, we analysed the text size in its English reference\footnote{Every translation from language A to B, we also translated the texts from language A to English, and same for language B, so we could make our analysis.} – language which also served as reference to evaluate the translation accuracy. To evaluate the translation accuracy, we used GLEU\footnote{This metric, mostly measures how accurate was a translation. It consists of taking the minimum recall and precision value of n-gram pairs between the compared texts; it is deeply explained in \cite{wu2016googles}.} (Google Bilingual Evaluation Metric) in two methods: \textbf{accumulated} (GLEU evaluation of the $t$-th resulting text with the original - means how is the $t$-th text when compared with the original) and \textbf{step by step} (GLEU evaluation of a step, from a language to the next one - measures how does each translation changes the text).

In addition, we propose another approach to evaluate semantic divergence (from sequential accuracy measurements) of translations. This approach is based on determining a function that best correlated the empirical values. To evaluate the correlation degree between the theoretical models and the empirical values, we used the loss function mixed Mean Square Error ($RMSE$), given by:
\begin{equation}
    RMSE = \sqrt{\frac{1}{N}\sum_{i=1}^N\left[\textrm{GLEU acc}(t) - \textrm{Theoretical}(t)\right]^2}
    \label{eq:rmse}
\end{equation}
In this equation $N$ is the chain size, GLEU acc$(t)$ refers to the GLEU accumulated value in the $t$-th translation. Theoretical$(t)$ is a mathematical function that measures the theoretical value for the same translation step – this function is what we wish to find by minimizing the RMSE. 

In Section \ref{subsec:resultadosC} we study how  the distance between languages impacts  each translation hop. For that, we use  \textit{common} and \textit{mixed} chains starting with text t1 and t3.
%Then we used a method in this article called GLEU \textbf{step-by-step} to calculate the accuracy between each translation hop. With that in hands, we average the values of each directed translation  that we obtain in the experiments to build a heatmap that describes the impact of the distance between languages and it's machine translation accuracy.

The following section describes our main results;  all relevant material can be found at
https://github.com/gecid-aia/babel/tree/article.

\section{Main Results}
\label{sec:resultados}

Section \ref{subsec:resultadosA} asks: how does sentences shrink along sequential translation?
In Section \ref{subsec:resultadosB}  an empirical law is introduced to capture the accuracy of sequential machine translation.
Section \ref{subsec:resultadosC} contains a brief discussion of the impact, in terms of accuracy, of the distance between the origin of languages.

\subsection{Text size decreases along Sequential Machine Translation}
\label{subsec:resultadosA}

While performing this first set of experiments, we identified some machine translation patterns also discussed in Ref.\ \cite{panter_2019}, where PANTER (2019) observes:

\begin{enumerate}
    \item \textbf{Fragmentation}: omission of connectors and pronouns, reducing readability.
    \item \textbf{Incoherent ordering}: Words assume an unconventional order as “day sunny” instead of “sunny day”, reducing text comprehension.
    \item \textbf{Literal translation instead of semantic dependency}: translation of terms isolated from the context, which in effect results in the semantic divergence of the sentence.
\end{enumerate}

Here are some excerpts from translation chains found in our research that illustrate   incoherence produced by these recurring inconsistencies:

\texttt{\\Step 0/284}\\
\texttt{>> t1:}\\
\texttt{“For him, each reading of any text will always provide a new resizing and understanding of that text. Metaphorically speaking, he positions himself before the texts as the bather of the river of Heráclito, in which it is impossible to enter twice because his course is constantly changing.”}\\

\texttt{Step 22/284}\\
\texttt{>> Maltese - English}\\
\texttt{“Regular reading each time explaining the contents and its meaning. On stage, he changed his approach to the Bible and putting in a rock on the corner of Geraeli.”}\\

\texttt{Step 78/284}\\
\texttt{>> Bosnian - English}\\
\texttt{“Please read first. He put the Bible on a stone with a scroll.”}\\

\texttt{Step 206/284}\\
\texttt{>> Czech - English}\\
\texttt{“Read and forget the Bible.”}\\

\texttt{Step 284/284}\\
\texttt{>> Pashto - English}\\
\texttt{“Read the Bible and forget it.”}

\begin{figure}[t]
    \floatbox[{\capbeside\thisfloatsetup{capbesideposition={right,top},capbesidewidth=5cm}}]{figure}[\FBwidth]
    {\caption{For both graphs, we show the text size variation along three randomized translation chains (rand1, rand2 and rand3). At the top, we started with t1 and t3 (initially 43 and 48 words, respectively), and after 284 translation hops, the mean text size was approximately 6 words, which is about 16\% of the initial size. At the bottom, we used t2 and t4 (a text with 296 and 258 words, respectively), in the case, after the same number of hops, the mean text size was only 7 words, which is about 2\% of the initial size.}
     \label{fig:size_t1_t2}}
    {\includegraphics[width=9cm]{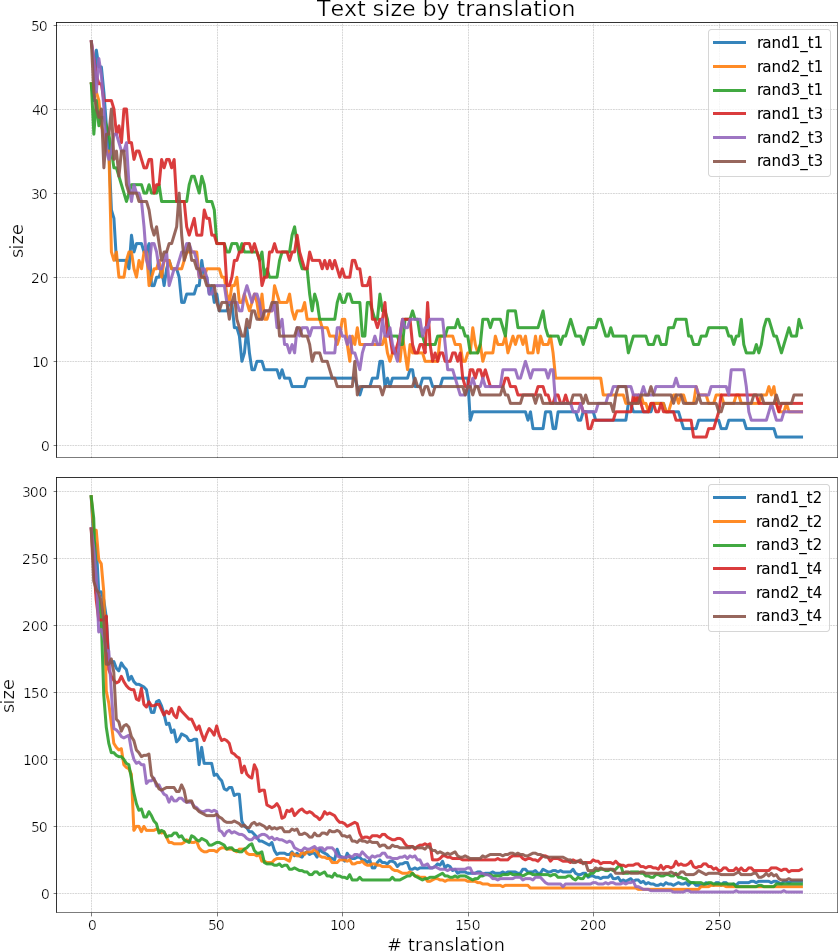}}
\end{figure}

Our study, however, not only confirmed the existence of drawbacks pointed by PANTER (2019), but also verified another pattern in machine translation: the fact that there is always a shrinkage of the input text, comparing not only beginning and the end of the translation chains but also each of the successive steps on average. In Figure \ref{fig:size_t1_t2}  we show that regardless of the size of the initial sentence, when applied to a randomized sequence of translations, the text size shrinks.

\begin{figure}[H]
    \floatbox[{\capbeside\thisfloatsetup{capbesideposition={left,top},capbesidewidth=5cm}}]{figure}[\FBwidth]
    {\caption{In the both graphs, the curves refer to text t1 and t3, with initially 43 and 48 words, respectively; and in both cases over translation chains with 284 hops. At the top, we observe the text size variation over the common chains (com1 and com2). At the bottom, the text size variation for the same texts although over the mixed translation chains (mix1 and mix2). As we can see, translation hops over common languages have a fewer impact when compared to translation hops over mixed languages; the average size of common translations after the 284 hops was 35.5 words, while in the mixed translations was 16.75.}
     \label{fig:size_rl}}
    {\includegraphics[width=9.5cm]{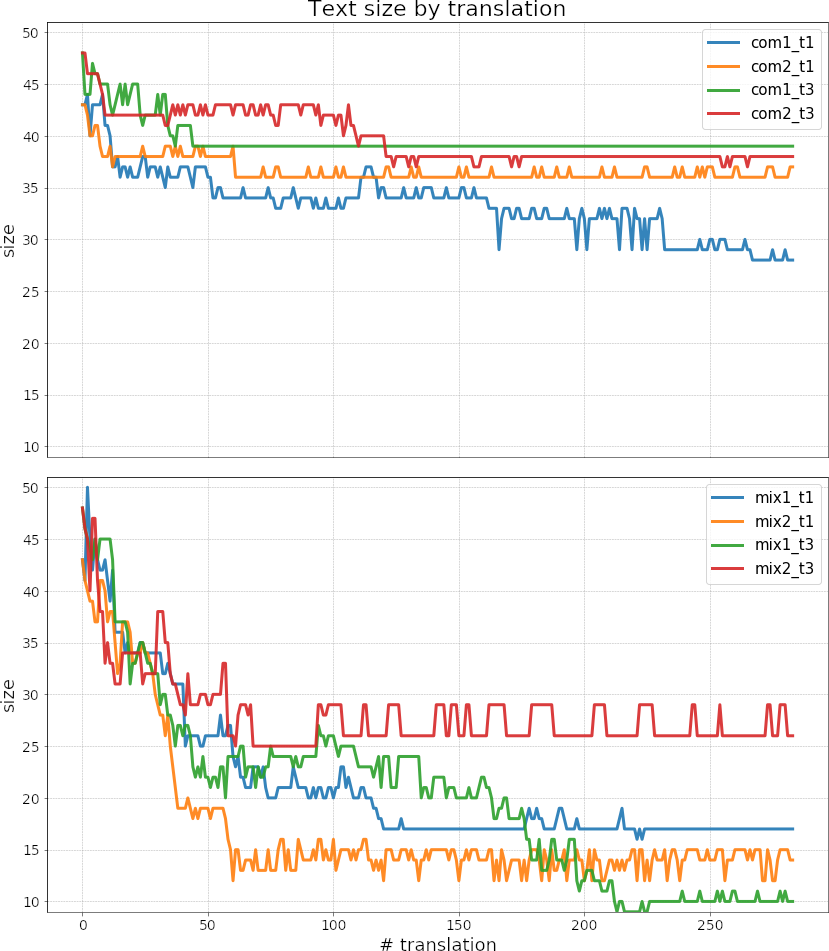}}
\end{figure}

In a second moment, we studied the impact of the translation when the same initial text is applied, by one side to a sequence of nearby languages, and by the other side, to distant languages. We found out that, the intensity of shrinkage is related to the distance between the languages, in other words, the nearest the languages of a set are, less shrinkage occurs in the text; and we verify the opposite to languages that are much more distant. In Figure \ref{fig:size_rl} we show how does the shrinkage differ in this two sets of experiments.

\subsection{The accuracy of sequential machine translation can be described as an empirical law}
\label{subsec:resultadosB}

Another recurring phenomenon we identified is the rather regular behavior of accuracy (GLEU accumulated method) in machine translation. We propose that it can be represented as a mathematical function ($AEL$ - Accuracy Empirical Law), a result that we think is novel. This law measures the GLEU accumulated accuracy along a chain of translation hops. In short, we suggest the following relationship between number of transition hop and accuracy:
\begin{equation}
    AEL(t) = (t+1)^{-\alpha}
    \label{eq:ael}
\end{equation}
%To determine this function, we start from a power law modeling that suggests the following behavior:
%\begin{equation}
%    p(t) = \beta\times t^{-\alpha}
%    \label{eq:potencia}
%\end{equation}
%In which $\alpha$ and $\beta$ are constants. 

In this expression we take the initial accuracy value (original text) to have value 1; $t$ refers to the $t$-th translation (the original text is presented at $t = 0$); $\alpha$ is a parameter that we call  {\em  semantic divergence factor}.

\begin{figure}[H]
    \floatbox[{\capbeside\thisfloatsetup{capbesideposition={right,top},capbesidewidth=5cm}}]{figure}[\FBwidth]
    {\caption{At the top, twelve curves that represent the GLEU accumulated of texts t1, t2, t3 and t4 applied to the translations chains rand1, rand2 and rand3. At the bottom, the red painted region refers to the confidence interval (using RMSE) of the six above curves, while the dashed line is the calculated AEL to the average of the curves. In result we have a semantic divergence factor $\alpha = 0.481$ and a RMSE = $0.033$.}
     \label{fig:gleuacc_b_all}}
    {\includegraphics[width=9.5cm]{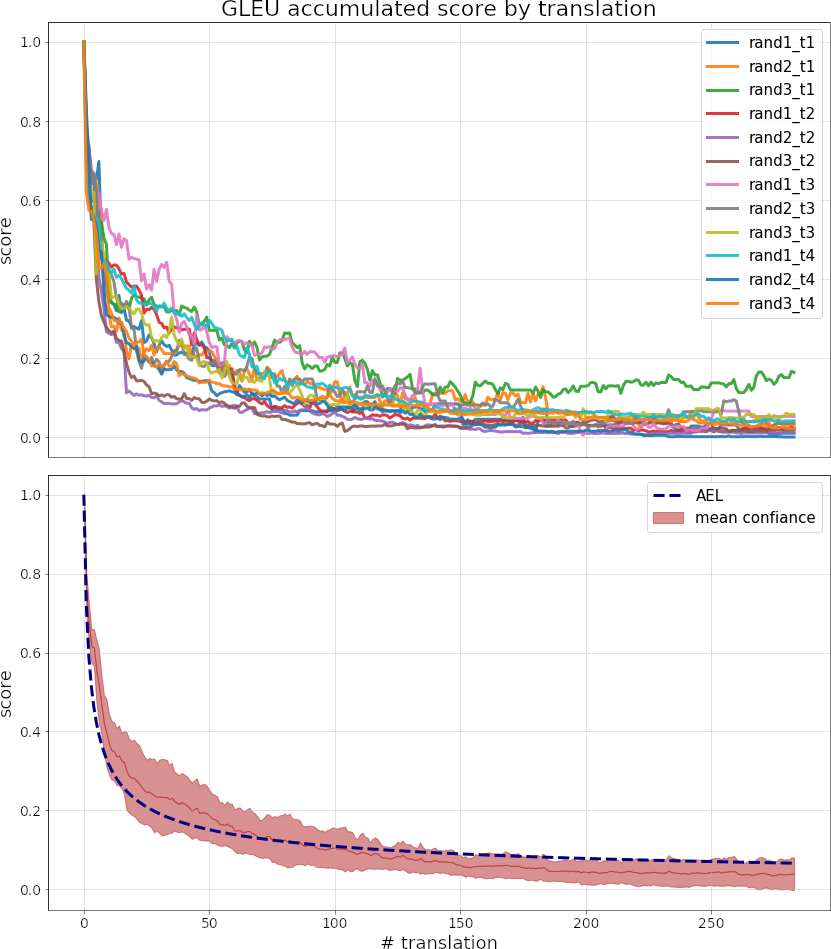}}
\end{figure}

To illustrate the AEL, consider the following  experiments. First we used the translation chains rand1, rand2 and rand3 with the initial texts t1, t2, t3 and t4. We thus got 12 different curves of GLEU accumulated that appear in Figure \ref{fig:gleuacc_b_all}. After measuring the empirical values, we minimized the RMSE (Eq. \ref{eq:rmse}), where GLEU acc is the curve for the average empirical values and the theoretical function is the AEL (Eq. \ref{eq:ael}). In the case, for each curve, the RMSE ranged from 0.02 to 0.04, which for the case justify a consistent modeling.

We then had the semantic divergence factor ($\alpha$) and the AEL curve that represents the average sequence machine translation along a set of 71 different languages in a random order, which is represented in Figure \ref{fig:gleuacc_b_all}.

\begin{figure}[H]
    \floatbox[{\capbeside\thisfloatsetup{capbesideposition={left,top},capbesidewidth=5cm}}]{figure}[\FBwidth]
    {\caption{The GLEU accumulated score curves for translation chains com1, com2, mix1 and mix2 using t1 and t3 as initial texts. It is remarkable that com1 and com2 translation chains decrease much less than mix1 and mix2 translation chains, suggesting that machine translation between nearby languages displays better performance than between distant languages.}
     \label{fig:gleuacc_rl}}
    {\includegraphics[width=9.5cm]{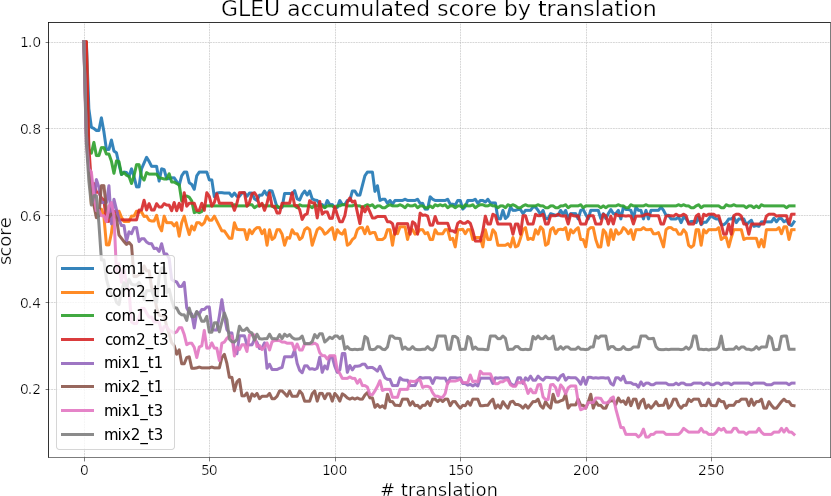}}
\end{figure}

We then used the translation chains com1 and com2 (nearby languages in family tree), and also mix1 and mix2 (more distant languages), to study the impact of the accuracy over sequential translation. In Figure \ref{fig:gleuacc_rl} we show how   each set of translation behaves along translation. One can see that while more distant languages (and fewer quantity of languages) behave similarly to the experiment with a sequence of 71 random languages, the nearby languages have a higher GLEU accumulated value along each translation hop. Text semantics is much better preserved using sequences with nearby languages than with distant languages.

\begin{figure}[H]
    \floatbox[{\capbeside\thisfloatsetup{capbesideposition={right,top},capbesidewidth=5cm}}]{figure}[\FBwidth]
    {\caption{At the top, we show  common-chains and at the bottom  
    mixed-chains. The red painted region is the confidence interval measured with RMSE and the purple dashed curve is the respective calculated AEL. For the \textit{common} group, the semantic divergence factor $\alpha = 0.110$ and the RMSE $= 0.040$, while for the \textit{mixed} group, $\alpha = 0.290$ and RMSE $= 0.022$. This means that, with nearby languages the calculated semantic divergence factor for this experiment is almost three times lower than for more distant languages.}
     \label{fig:gleuacc_rl_2}}
    {\includegraphics[width=9.5cm]{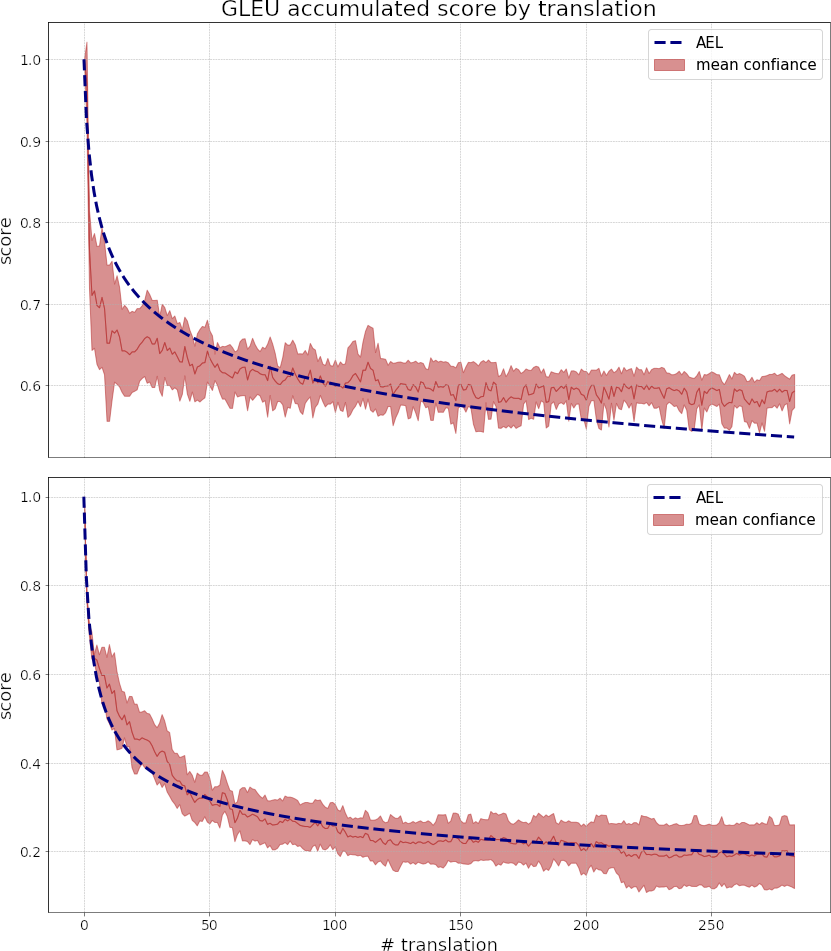}}
\end{figure}

We then evaluated the AEL for common-chains and mixed-chains, in a way to calculates the semantic divergence factor over different distance degrees of languages. As we expected, the semantic divergence factor for nearby languages is lower than for more distant languages, see Figure \ref{fig:gleuacc_rl_2}.

From this set of experiments we not only identify a monotonic decrease on average, but also the fact that we could model it using the $AEL$ approach. As the semantic divergence factors for chains with only seven languages are lower than using 71 languages, it seems that greater linguistic diversity leads to a lower accuracy by sequential machine translation. 

\subsection{Visualization of translation accuracy}
\label{subsec:resultadosC}

\begin{figure}[H]
    \caption{In this set of heatmaps, we visualize the average GLEU score with the step-by-step method; each square represent the translation accuracy from the language listed in the row to the language listed in the column. In this case, the hotter the color, the better the average translation accuracy; and it is notable that near languages indeed have a better translation between each other (\textbf{a} - com1 and \textbf{b} - com2) than languages that are more distant (\textbf{c} - mix1 and \textbf{d} - mix2). In the color bar at the bottom, we signalized the average value of each heatmap (excluding translation from language A to A — which does not occur); the values for \textbf{a}, \textbf{b}, \textbf{c} and \textbf{d} are, respectively, 0.966, 0.949, 0.899 and 0.902.}
    \label{fig:gleupap1}
    \includegraphics[width=15cm]{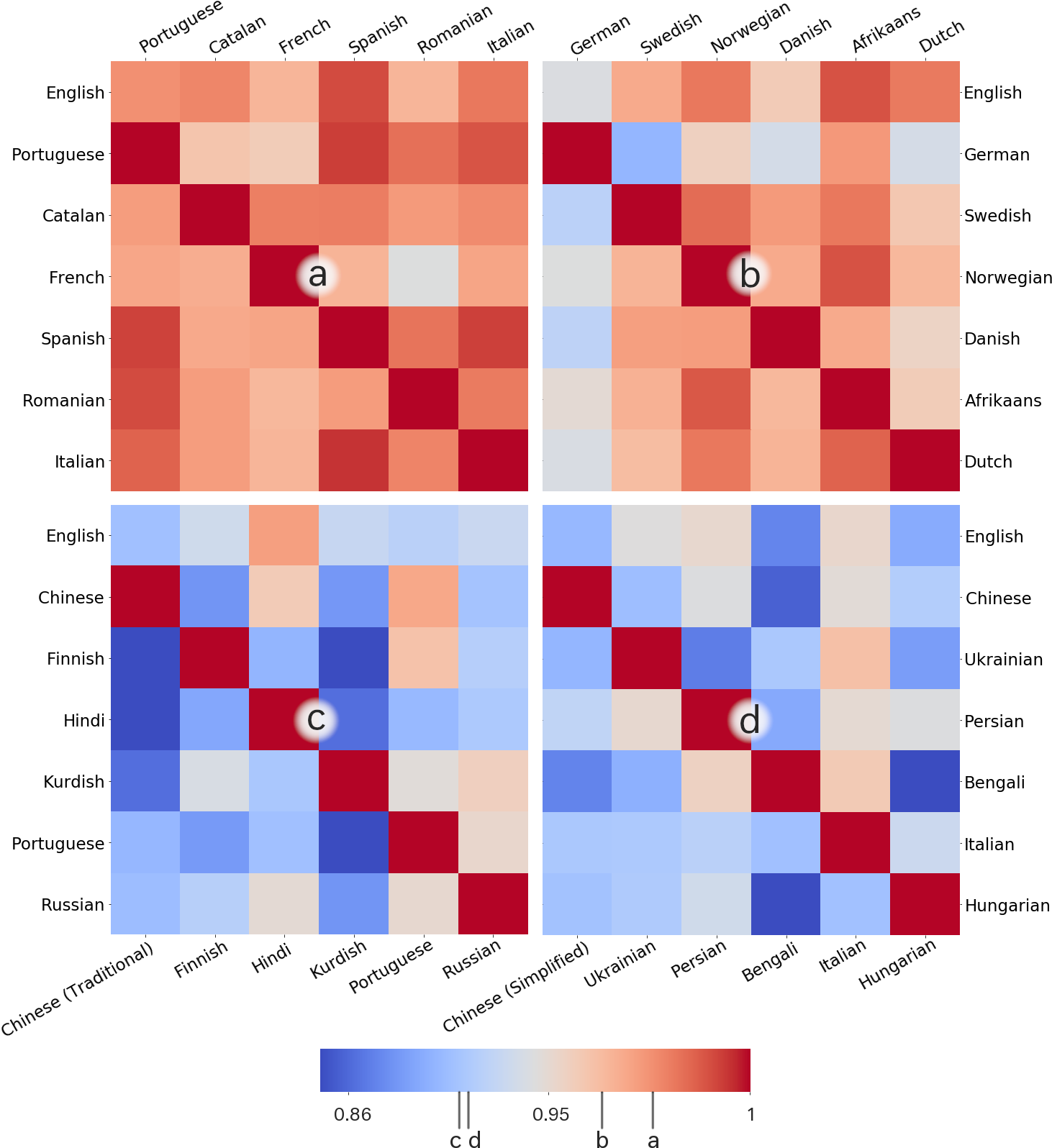}
\end{figure}

Now consider analyzing translation accuracy through a heatmap, where we use the GLEU step-by-step method — that is, calculate the GLEU between two adjacent texts produced in the translation chain.
In Figure \ref{fig:gleupap1} we have a heatmap representation of 144 directed translation pairs. On average, we measured seven times each directed translation accuracy using GLEU step-by-step from language A to B. The figure is divided into four sub-images, \textbf{a} and \textbf{b} are the pairs for the common-chains; \textbf{c} and \textbf{d} are the pairs for the mixed-chains. In the map, we confirm that translation accuracy using nearby languages is higher (hot colors) than translation accuracy using more distant languages (cold colors).

Analyzing a smaller subset of languages is a way of understanding in more detail the behavior of specific moments in translation chains. We leave open here the possibility to make a heatmap for each of the 103 language combinations in Google Translate – then it would be possible to construct an overview of translation behavior among all language pairs of this system.
 
\section{Conclusion}
\label{sec:conclusoes}

Our experiments show that sequential translation progressively shrinks translated texts, even for languages that are close to each other — that is, in the same language family. Sequences that use 71-languages  shorten the text of the sequences as compared to just 7 languages, when going through 284 translation hops.

We also found, perhaps less surprisingly, that translation sequences with languages in the same family maintain higher accuracy  than translation sequences with distant languages (distinct families).
%, as well as sequences with a much larger number of languages (71).

And we also detected an empirical curve ($AEL$) that can predict surprisingly well the accuracy behavior of the GLEU accumulated method. This is our main contribution as it suggests that one can:
\begin{itemize}
    \item Predict accuracy even before actual translation is performed. Also, one can identify sequences of translations that are better than a direct translation (say instead of going from A to B, try A to C to D to B). Moreover, one can identify worst performing language sequences (high semantic divergence factor) to improve specific steps with translation chains.
    \item Understand how languages relate to each other regarding accuracy --– which sets of languages behave better then others, thus leading to a new kind of language ordering, no longer from its family ancestry, but from its relational accuracy.
\end{itemize}
While the first bullet lists a number of practical contributions of our study, the second bullet may have far-reaching consequences within linguistics. 

Overall, this study describes a possible way to investigate properties of black box techniques in a holistic manner. We intend to continue
this investigation with additional language features and families.
%\footnote{In the book The Black Box Society \cite{pasquale_2015}, Frank Pasquale quotes: “Black box may refer to a system whose workings are mysterious; we can observe its inputs and outputs, but we cannot tell how one becomes the other.”} employed in machine translation \cite{mims_2017}. 

\section*{Acknowledgements}

This research would not be possible without the collaboration of Pedro Barbosa, Paloma F. Victor (for supporting the first author in the analysis of the results), members of Art and Artificial Intelligence Group (GAIA) / Critical Experiences in Digital Infrastructure Group (GECID) / C4AI, employees of Inova USP, and PIBIQ (Institutional Scientific Initiation Scholarship Program).

The second author (Bruno Moreschi) is partially supported by Research and Collaboration Awards on the Histories of AI: A Genealogy of Power - Sawyer Seminar, Cambridge University; and Center for Arts, Design \& Social Research.

The third author (Fabio Cozman) is partially supported by the CNPq grant 312180/2018-7 (PQ). This work has been supported in part by 
Fundação de Amparo à Pesquisa 
do Estado de São Paulo (FAPESP), grants 2015/21880-4, 2016/18841-0, 2019/07665-4, 
 and in part by the Coordenação de Aperfeiçoamento de Pessoal de Nível Superior (CAPES) – finance code 001.

\bibliographystyle{IEEEtran}  
\bibliography{references}  %%% Remove comment to use the external .bib file (using bibtex).


\end{document}